# Bounded Approximate Symbolic Dynamic Programming for Hybrid MDPs

**Luis Gustavo Rocha Vianna**
University of Sao Paulo
Sao Paulo, Brazil
ludygrv@ime.usp.br

**Scott Sanner**
NICTA & ANU
Canberra, Australia
ssanner@nicta.com.au

**Leliane Nunes de Barros**
University of Sao Paulo
Sao Paulo, Brazil
leliane@ime.usp.br

## Abstract

Recent advances in symbolic dynamic programming (SDP) combined with the extended algebraic decision diagram (XADD) data structure have provided exact solutions for mixed discrete and continuous (hybrid) MDPs with piecewise linear dynamics and continuous actions. Since XADD-based exact solutions may grow intractably large for many problems, we propose a bounded error compression technique for XADDs that involves the solution of a constrained bilinear saddle point problem. Fortuitously, we show that given the special structure of this problem, it can be expressed as a bilevel linear programming problem and solved to optimality in finite time via constraint generation, despite having an infinite set of constraints. This solution permits the use of efficient linear program solvers for XADD compression and enables a novel class of bounded approximate SDP algorithms for hybrid MDPs that empirically offers order-of-magnitude speedups over the exact solution in exchange for a small approximation error.

## 1 Introduction

Many real-world sequential-decision making problems involving resources, time, or spatial configurations naturally use continuous variables in both their state and action representation and can be modeled as Hybrid Markov Decision Processes (HMDPs). While HMDPs have been studied extensively in the AI literature [4; 7; 10; 9; 11; 12], only recently have symbolic dynamic programming (SDP) [14; 17] techniques been introduced to enable the exact solution of multivariate HMDPs with continuous actions and arbitrary piecewise linear dynamics and rewards.

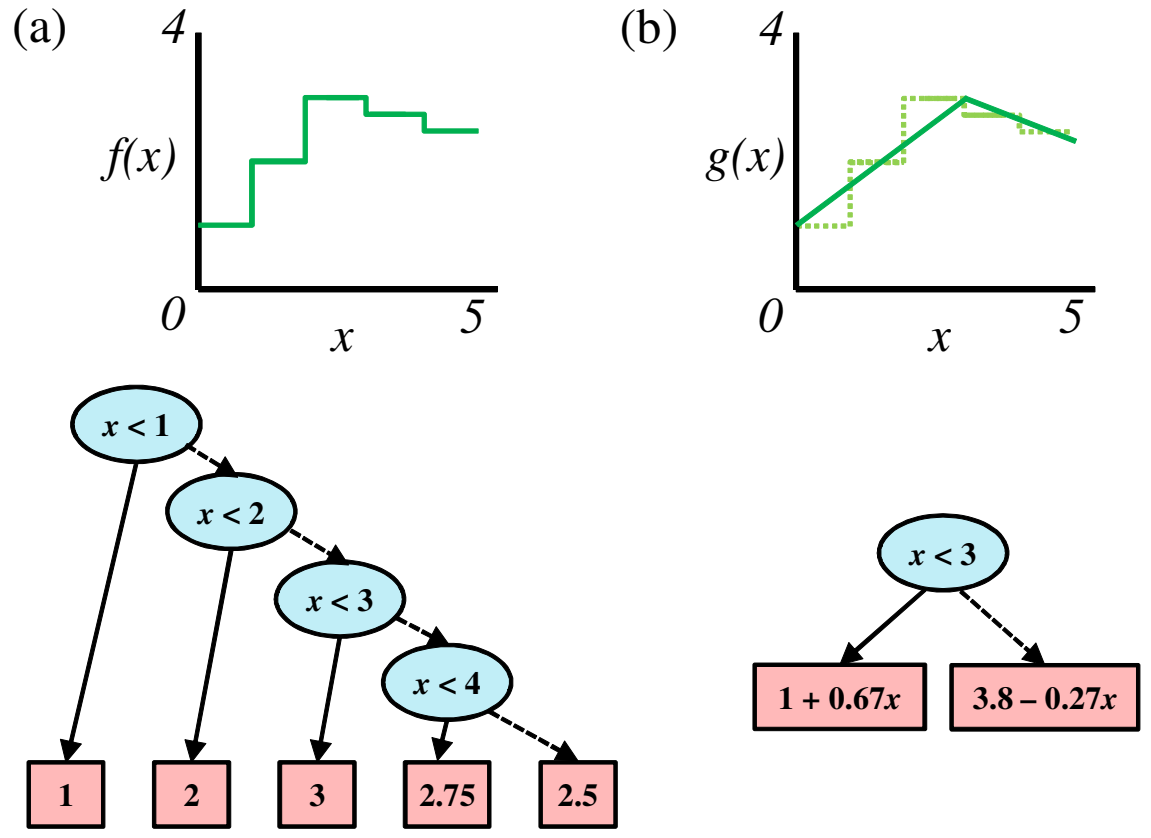

Figure 1: (a) A function $f(x)$ for $x \in [0, 5]$ and its XADD representation (solid branch is true, dotted branch is false); (b) A compressed XADD approximation $g(x)$ of $f(x)$. While these simple XADDs are trees, XADDs are more generally directed acyclic graphs as we show later.

What has proved crucial in this SDP solution of piecewise linear HMDPs is the use of the XADD data structure representation of functions like the simple examples shown in Figure 1(a,b) that allows the HMDP value function to be represented compactly and SDP operations to be computed efficiently. In brief, an XADD is simply an extension of the algebraic decision diagram (ADD) [1] to continuous variables where decisions may be boolean variable tests or inequalities of continuous expressions and leaves may be continuous expressions; XADDs are evaluated from root to leaf like decision trees. Following the SDP work of [17] for HMDPs with continuous actions that we extend, we restrict XADDs to have linear decisions and leaves.

While XADDs have enabled SDP solutions to HMDPs that would not be otherwise possible with more naïve representations of piecewise functions, XADDs still have limitations — for some problems the HMDP solution (represented by a value function) simply has many distinct pieces and does not admit a more compact *exact* XADD representation, e.g, Figure 1(a). However, motivated by previous approximation work in discrete factored MDPs using ADD approximation [16], we pose the question of whether there exists a method for compressing an XADD in exchange for some bounded approximation error. As a hint of the solution, we note that Figure 1(a) can be approximated by 1(b) which is more compact and induces relatively small error.

But how do we find such a compressed XADD? In the simpler case of ADDs [16], this approximation process was straightforward: leaves with nearby constant values are averaged and merged, leading to bottom-up compaction of the ADD. In the XADD, if we wish to take a similar approach, we see that the problem is more complex since it is not clear (1) which leaves to merge, or (2) how to find the best approximation of these leaves that minimizes the error over the *constrained, continuous* space where each leaf is valid. Indeed, as Figure 1(a,b) demonstrates, the answer is *not* given simply by averaging leaves since the average of constant leaves in (a) could never produce the linear function in the leaves of (b). Hence, we wish to answer questions (1) and (2) to produce a *bounded and low-error approximation over the entire continuous function domain* as given in Figure 1(b).

To answer these questions, we propose a bounded error compression technique for linear XADDs that involves the solution of a constrained bilinear saddle point problem. Fortuitously, we show that given the special structure of this problem, it can be expressed as a bilevel linear programming problem. While the second-level optimization problem in this bilevel program implicitly represents an infinite number of constraints, we show that a constraint generation approach for this second stage allows the first stage to terminate at optimality after generating only a finite number of constraints. This solution permits the use of efficient linear program solvers for XADD compression and enables a novel class of bounded approximate SDP algorithms for hybrid MDPs. Empirically we demonstrate that this approach to XADD compression offers order-of-magnitude speedups over the exact solution in exchange for a small approximation error, thus vastly expanding the range of HMDPs for which solutions with strong error guarantees are possible.

## 2 Extended Algebraic Decision Diagrams (XADDs)

We begin with a brief introduction to the extended algebraic decision diagram (XADD), then in Section 3 we contribute approximation techniques for XADDs. In Section 4, we will show how this approximation can be used in a bounded approximate symbolic dynamic programming algorithm for hybrid MDPs.

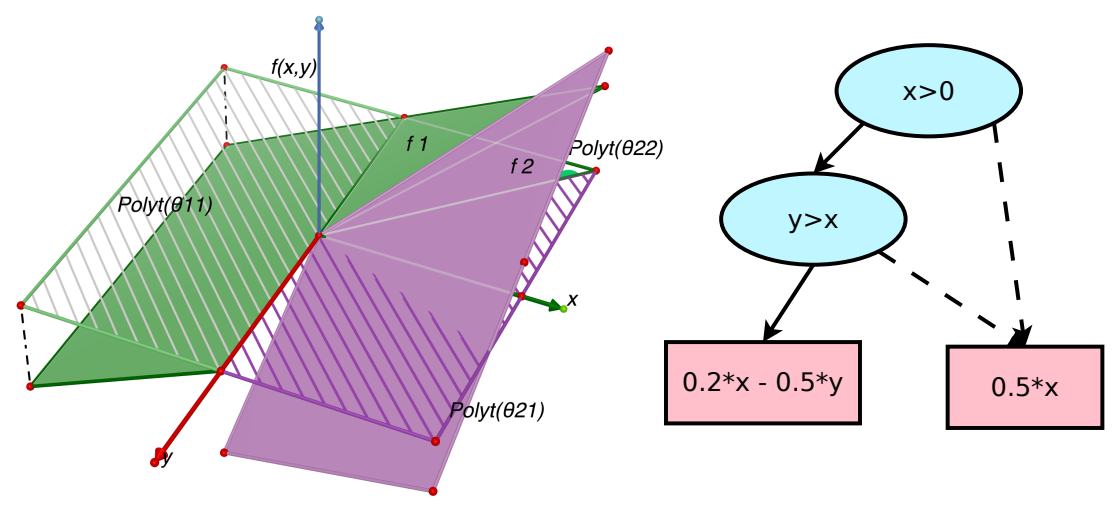

$$f(x,y) = \begin{cases} \phi_1 : f_1 \\ \phi_2 : f_2 \end{cases} \quad \begin{aligned} \phi_1 &= \theta_{11} \vee \theta_{12} \\ \theta_{11} &= x < 0 \\ \theta_{12} &= x > 0 \wedge x < -y \\ f_1 &= \frac{x}{2} \end{aligned} \quad \begin{aligned} \phi_2 &= \theta_{21} \\ \theta_{21} &= x > 0 \wedge y > x \\ f_2 &= \frac{x}{5} - \frac{y}{2} \end{aligned}$$

Figure 2: Example of piecewise linear function in case and XADD form: *(top left)* plot of function $f(x,y)$; *(top right)* XADD representing $f(x,y)$; *(bottom)* case semantics for $f(x,y)$ demonstrating notation used in this work.

### 2.1 Case Semantics of XADDs

An XADD is a function represented by a directed acyclic graph having a fixed ordering of decision tests from root to leaf. For example, Figure 2(top left) shows the plot of a piecewise function and 2(top right) its XADD representation. Underlying this XADD is a simple piecewise linear function that we can represent semantically in case form. Specifically, given a domain of boolean and continuous variables $(\boldsymbol{b}^T, \boldsymbol{x}^T) = (b_1, \ldots, b_n, x_1, \ldots, x_m)$, where $b_i \in \{0,1\}$ $(1 \le i \le n)$ and $x_j \in [x_j^{\min}, x_j^{\max}]$ $(1 \le j \le m)$ for $x_j^{\min}, x_j^{\max} \in \mathbb{R}$ and $x_j^{\max} > x_j^{\min}$, a case statement representing an XADD with linear decisions and leaf expressions takes the following piecewise linear form

$$f(\boldsymbol{b}, \boldsymbol{x}) = \begin{cases} \phi_1(\boldsymbol{b}, \boldsymbol{x}) : & f_1(\boldsymbol{x}) \\ \vdots & \vdots \\ \phi_k(\boldsymbol{b}, \boldsymbol{x}) : & f_k(\boldsymbol{x}) \end{cases} . \quad (1)$$

Here the $f_i$ are linear expressions over $\boldsymbol{x}$ and the $\phi_i$ are logical formulae defined over the domain $(\boldsymbol{b}^T, \boldsymbol{x}^T)$ that can include arbitrary logical $(\wedge, \vee, \neg)$ combinations of (i) boolean variables and (ii) linear inequalities over $\boldsymbol{x}$.

In the XADD example of Figure 2, *every leaf* represents a case value $f_i$ and *every path from root to leaf* represents a conjunction of decision constraints. The disjunction of all path constraints leading to leaf $f_i$ corresponds to a case partition $\langle \phi_i(\boldsymbol{b}, \boldsymbol{x}) : f_i(\boldsymbol{x}) \rangle$. Clearly, all case partitions derived from an XADD must be mutually disjoint and exhaustive of the domain $(\boldsymbol{b}^T, \boldsymbol{x}^T)$, hence XADDs represent well-defined functions.

Since $\phi_i$ can be written in disjunctive normal form (DNF), i.e., $\phi_i \equiv \bigvee_{j=0}^{n_i} \theta_{ij}$ where $\theta_{ij}$ represents a conjunction of linear constraints over $\boldsymbol{x}$ and a (partial) truth assignment to $\boldsymbol{b}$ corresponding to the $j$th path from the XADD root to leaf $f_i$, we observe that every $\theta_{ij}$ contains a bounded convex linear polytope ($\boldsymbol{x}$ is finitely bounded in all dimensions as initially defined). We formally define the set of all convex linear polytopes contained in $\phi_i$ as $C(\phi_i) = \{Polytope(\theta_{ij})\}_j$, where *Polytope* extracts the subset of linear constraints from $\theta_{ij}$. Figure 2(top left) illustrates the different polytopes in the XADD of 2(top right) with corresponding case notation in 2(bottom).

### 2.2 XADD Operations

XADDs are important not only because they compactly represent piecewise functions that arise in the forthcoming solution of hybrid MDPs, but also because operations on XADDs can efficiently exploit their structure. XADDs extend algebraic decision diagrams (ADDs) [1] and thus inherit most unary and binary ADD operations such as addition $\oplus$ and multiplication $\otimes$. While the addition of two linear piecewise functions represented by XADDs remains linear, in general their product may not (i.e., the values may be quadratic); however, for the purposes of symbolic dynamic programming later, we remark that we only ever need to multiply piecewise constant functions by piecewise linear functions represented as XADDs, thus yielding a piecewise linear result.

Some XADD operations do require extensions over the ADD, e.g., the binary max operation represented here in case form:

$$\max\left( \begin{cases} \phi_1 : f_1 \\ \phi_2 : f_2 \end{cases}, \begin{cases} \psi_1 : g_1 \\ \psi_2 : g_2 \end{cases} \right) = \begin{cases} \phi_1 \wedge \psi_1 \wedge f_1 > g_1 : f_1 \\ \phi_1 \wedge \psi_1 \wedge f_1 \le g_1 : g_1 \\ \phi_1 \wedge \psi_2 \wedge f_1 > g_2 : f_1 \\ \phi_1 \wedge \psi_2 \wedge f_1 \le g_2 : g_2 \\ \vdots \qquad \vdots \end{cases}$$

While the max of two linear piecewise functions represented as XADDs remains in linear case form, we note that unlike ADDs which prohibit continuous variables $\boldsymbol{x}$ and *only* have a *fixed* set of boolean decision tests $\boldsymbol{b}$, an XADD may need to create *new decision tests* for the linear inequalities $\{f_1 > g_1, f_1 > g_2, f_2 > g_1, f_2 > g_2\}$ over $\boldsymbol{x}$ as a result of operations like max.

Additional XADD operations such as symbolic substitution, continuous (action) parameter maximization, and integration required for the solution of hybrid MDPs have all been defined previously [14; 17] and we refer the reader to those works for details.

## 3 Bounded XADD Approximation

In this section, we present the main novel contribution of our paper for approximating XADDs within a fixed error bound. Since the point of XADD approximation is to shrink its size, we refer to our method of approximation as XADD Compression (`XADDComp`).

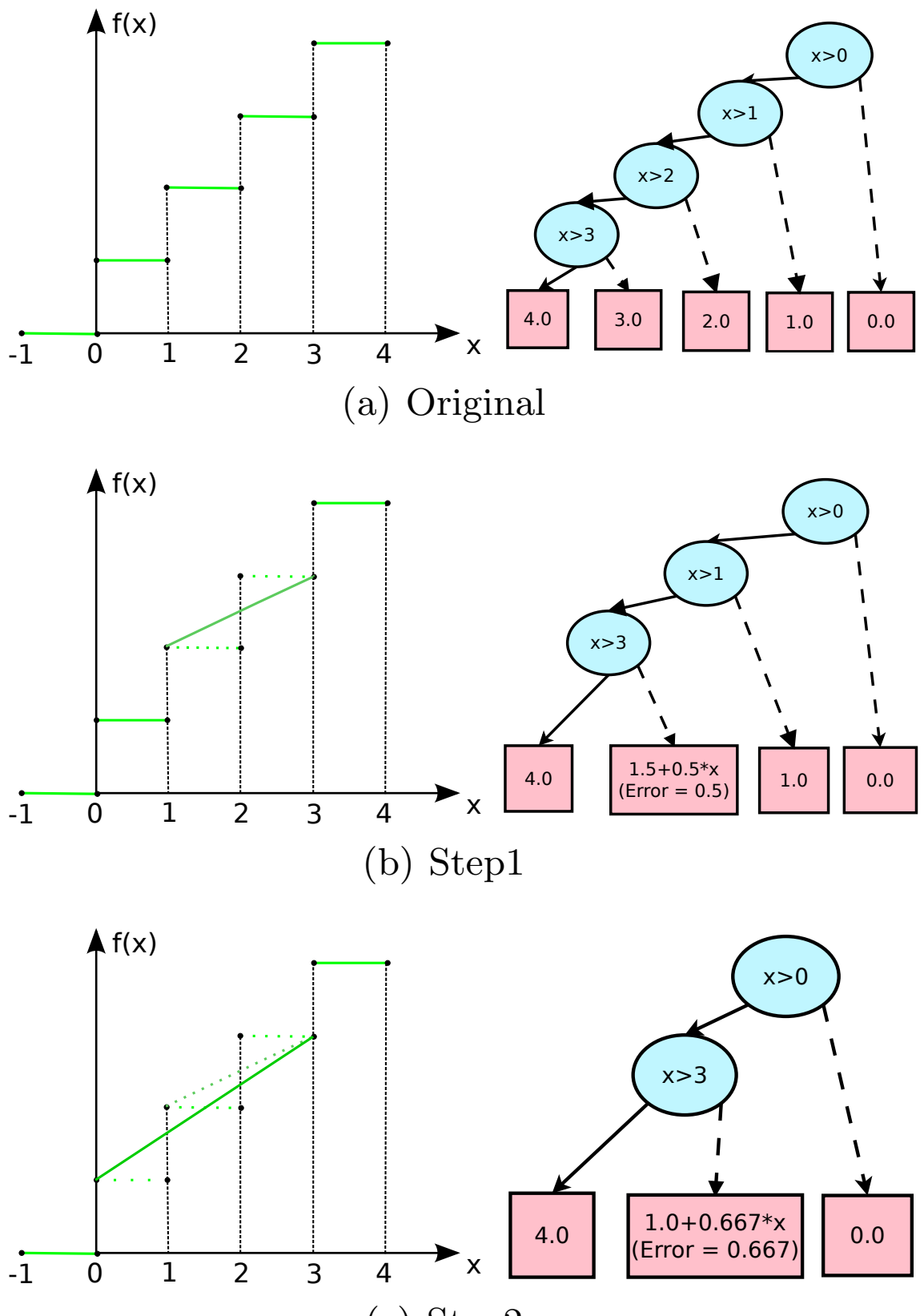

(a) Original

(b) Step1

(c) Step2

Figure 3: Successive pair merging XADD compression for a simple 1D example. At each step two nodes are chosen for merging, the best approximating hyperplane is determined according to Section 3.2, and if the accumulated error is within required bounds, the leaves are merged and internal XADD structure simplified to remove unneeded decisions.

Following previous work on ADD compression [16], we note that decision diagrams should be compressed from the bottom up — merging leaves causes simplifications to ripple upwards through a decision diagram removing vacuous decisions and shrinking the decision diagram. For example, after merging leaves in Figure 1(a), we note that the only remaining decision in 1(b) is $x < 3$. Hence, we focus on a leaf merging approach to `XADDComp`, which poses two questions: (1) what leaves do we merge? And (2) how do we find the best approximation of merged leaves? We answer these two questions in the following subsections.

### 3.1 Successive Leaf Merging

Since it would be combinatorially prohibitive to examine all possible leaf merges in an XADD, our `XADDComp` approximation approach in Algorithm 1 uses a systematic search strategy of successive pairwise merging of leaves. The idea is simple and is illustrated in Figure 3. The bounded error property is guaranteed by accumu-

**Algorithm 1**: XADDComp(XADD $X$, $\epsilon$) $\longrightarrow (\hat{X}, \hat{\epsilon})$

$\hat{\epsilon} \leftarrow 0$ *// The max amount of error used so far*
$\hat{X} \leftarrow X$ *// The approximated XADD*
$Open := \{L_i\} = \{\langle \phi_i, f_i \rangle\} \in \hat{X}$ *// cases in* $\hat{X}$
**while** $Open \neq \emptyset$ **do**
  $L_1 := Open.pop()$
  **for** $L_2 \in Open$ **do**
    *// Merge and track accumulated error*
    $(f^*, \tilde{\epsilon}) :=$ PairLeafApp$(L_1, L_2)$ *// Sec 3.2*
    $f^*.error := \tilde{\epsilon} + \max(f_1.error, f_2.error)$
    *// Keep merge if within error bounds*
    **if** $f^*.error < \epsilon$ **then**
      $\hat{\epsilon} := \max(\hat{\epsilon}, f^*.error)$
      $Open.remove(L_2)$
      *// Replace leaves in* $\hat{X}$ *and simplify*
      $\hat{X}.f_1 := f^*,\ \hat{X}.f_2 := f^*$
      $L_1 := \langle \phi_1 \vee \phi_2, f^* \rangle$ *// Keep merging* $L_1$
**return** $(\hat{X}, \hat{\epsilon})$ *// Comp. XADD and error used*

lating the amount of error "used" in every merged leaf and avoiding any merges that exceed a maximum error threshold $\epsilon$. However, we have not yet defined how to find the lowest error approximation of a pair of leaves in PairLeafApp, which we provide next.

### 3.2 Pairwise Leaf Approximation

In pairwise leaf merging, we must address the following fundamental problem: given two XADD leaves represented by their case partitions $L_1 = \langle f_1, \phi_1 \rangle$ and $L_2 = \langle f_2, \phi_2 \rangle$, our goal is to determine the best linear case approximation of $L_1$ and $L_2$. As it must represent $L_1$ and $L_2$, the solution must be defined in both regions and is therefore of the form $L^* = \langle f^*, \phi_1 \vee \phi_2 \rangle$. Since we restrict to linear XADDs, then $f_1 = \boldsymbol{c_1}^T(\boldsymbol{x}^T, 1)^T$, $f_2 = \boldsymbol{c_2}^T(\boldsymbol{x}^T, 1)^T$ and $f^* = \boldsymbol{c^*}^T(\boldsymbol{x}^T, 1)^T$ (assuming $\boldsymbol{c_1}, \boldsymbol{c_2}, \boldsymbol{c^*} \in \mathbb{R}^{m+1}$ where $\boldsymbol{x} \in \mathbb{R}^m$). Thus, our task reduces to one of finding the optimal weight vector $\boldsymbol{c^*}$ which minimizes approximation error given by the following bilinear saddle point optimization problem:

$$\min_{\boldsymbol{c}^*} \max_{i \in \{1,2\}} \max_{\boldsymbol{x} \in C(\phi_i)} \left| \underbrace{\boldsymbol{c_i}^T \begin{bmatrix} \boldsymbol{x} \\ 1 \end{bmatrix}}_{f_i} - \underbrace{\boldsymbol{c^*}^T \begin{bmatrix} \boldsymbol{x} \\ 1 \end{bmatrix}}_{f^*} \right| \qquad (2)$$

This is bilinear due to the inner product of $\boldsymbol{c^*}$ with $\boldsymbol{x}$.

To better understand the structure of this bilinear saddle point problem, we refer to Figure 4, which shows on the left, two functions $f_1$ and $f_2$ and the respective single polytope regions $C(\phi_1)$ and $C(\phi_2)$ where $f_1$ and $f_2$ are respectively valid. On the right, we show a proposed approximating hyperplane $f$ within the regions $C(\phi_1)$ and $C(\phi_2)$. Clearly we want to choose the $f^* = f$ that minimizes the absolute difference between $f$ and $f_1, f_2$ within their respective polytopes. On account of this perspective and recalling that $C(\phi_i) = \{Polytope(\theta_{ij})\}_j$, we can rewrite (2) as the following bi-level linear optimization problem:[1]

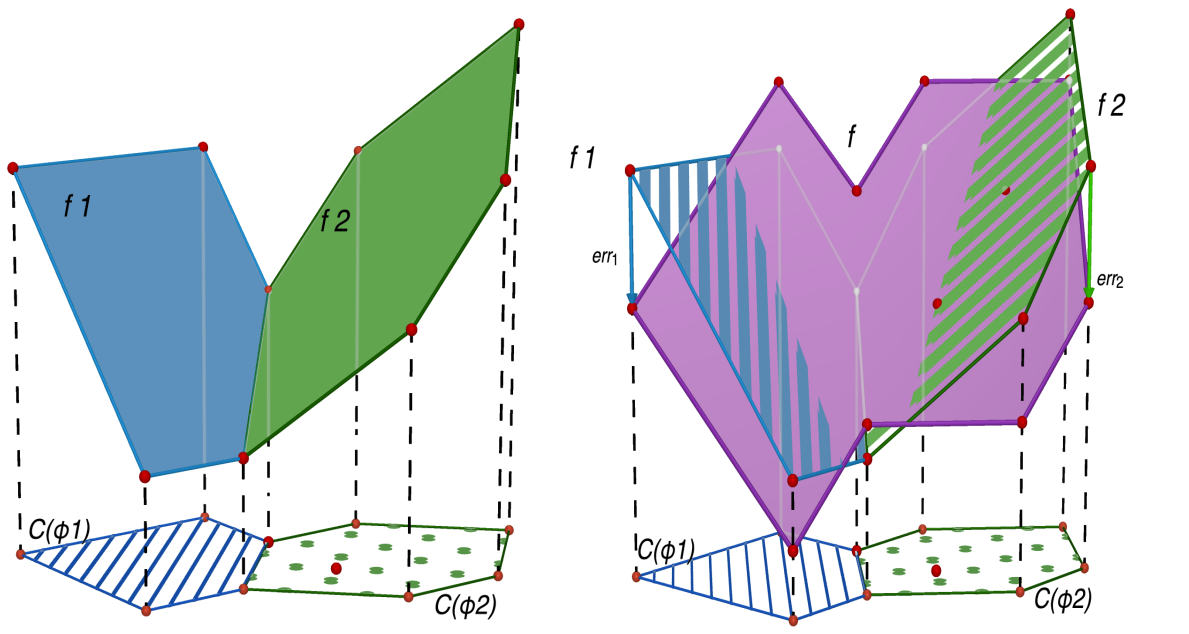

Figure 4: Illustration of the pairwise leaf approximation problem: *(left)* the original linear leaf functions $f_1$ and $f_2$ in their respective (single) polytope regions $\phi_1$ and $\phi_2$; *(right)* a linear approximation $f$ overlaid on $f_1$ and $f_2$ in their regions showing errors at the polytope vertices.

$$\min_{\boldsymbol{c}^*, \epsilon} \epsilon \qquad (3)$$
$$s.t.\ \epsilon \geq \begin{pmatrix} \max\limits_{\boldsymbol{x}} & \left| \boldsymbol{c_i}^T \begin{bmatrix} \boldsymbol{x} \\ 1 \end{bmatrix} - \boldsymbol{c^*}^T \begin{bmatrix} \boldsymbol{x} \\ 1 \end{bmatrix} \right| \\ s.t. & \boldsymbol{x} \in Polytope(\theta_{ij}) \end{pmatrix} ; \forall i \in \{1{,}2\}, \forall \theta_{ij}$$

While it may seem we have made little progress with this rewrite of (2) — this still appears to be a difficult optimization problem, we can make an important insight that allows us to remove the second stage of optimization altogether. While implicitly it appears that the second stage would correspond to an infinite number of constraints — one for each $\boldsymbol{x} \in Polytope(\theta_{ij})$, we return to Figure 4. Since each of $f$, $f_1$, and $f_2$ are all linear and $C(\phi_1), C(\phi_2)$ represent (unions of) linear convex polytopes, we know that the *maximum difference* between $f$ and $f_1, f_2$ must occur at the *vertices* of the respective polytope regions. Thus, denoting $\boldsymbol{x_{ij}^k}$ ($k \in \{1 \ldots N_{ij}\}$) as a vertex of the linear convex polytope defined by $\theta_{ij}$, we can obtain a *linear program* version of (2) with a *finite* number of constraints at the vertices $\boldsymbol{x_{ij}^k}$ of all polytopes:

$$\min_{\boldsymbol{c}^*, \epsilon} \epsilon \qquad (4)$$
$$s.t.\ \epsilon \geq \left| \boldsymbol{c_i}^T \begin{bmatrix} \boldsymbol{x_{ij}^k} \\ 1 \end{bmatrix} - \boldsymbol{c^*}^T \begin{bmatrix} \boldsymbol{x_{ij}^k} \\ 1 \end{bmatrix} \right| ; \quad \begin{array}{l} \forall i \in \{1{,}2\}, \forall \theta_{ij}, \\ \forall k \in \{1 \ldots N_{ij}\} \end{array}$$

[1] To obtain a true bi-level *linear* program, we need *two* separate second stage constraints to encode that $\epsilon$ is larger than each side of the absolute value (the argument of the absolute value and its negation), but this is a straightforward absolute value expansion in a linear program that we will consider implicit to avoid notational clutter.

**Algorithm 2**: $\texttt{PairLeafApp}(L_1, L_2) \longrightarrow (\boldsymbol{c}^*, \epsilon)$

1 $\boldsymbol{c^*} := \mathbf{0}$ // *Arbitrarily initialize* $\boldsymbol{c^*}$
2 $\epsilon^* := \infty$ // *Initialize to invalid error*
3 $C := \emptyset$ // *Start with an empty constraint set*
4 // *Generate max error vertex constraints for* $\boldsymbol{c^*}$
5 **for** $i \in \{1, 2\}, \theta_{ij} \in C(\phi_i)$ **do**
6
$$\boldsymbol{x^k_{ij}}_+ := \arg\max_{\boldsymbol{x}} \left( \boldsymbol{c_i}^T \begin{bmatrix} \boldsymbol{x} \\ 1 \end{bmatrix} - \boldsymbol{c^*}^T \begin{bmatrix} \boldsymbol{x} \\ 1 \end{bmatrix} \right) \quad \text{s.t. } \boldsymbol{x} \in \mathit{Polytope}(\theta_{ij})$$
$$\boldsymbol{x^k_{ij}}_- := \arg\max_{\boldsymbol{x}} \left( \boldsymbol{c^*}^T \begin{bmatrix} \boldsymbol{x} \\ 1 \end{bmatrix} - \boldsymbol{c_i}^T \begin{bmatrix} \boldsymbol{x} \\ 1 \end{bmatrix} \right) \quad \text{s.t. } \boldsymbol{x} \in \mathit{Polytope}(\theta_{ij})$$
$$C := C \cup \{\epsilon > \boldsymbol{c_i}^T (\boldsymbol{x^k_{ij}}_+^T, 1)^T - \boldsymbol{c^*}^T (\boldsymbol{x^k_{ij}}_+^T, 1)^T\}$$
7 $C := C \cup \{\epsilon > \boldsymbol{c^*}^T (\boldsymbol{x^k_{ij}}_-^T, 1)^T - \boldsymbol{c_i}^T (\boldsymbol{x^k_{ij}}_-^T, 1)^T\}$
8 // *Re-solve LP with augmented constraint set*
9 $(\boldsymbol{c^*}, \epsilon^*_{new}) := \arg\min_{\boldsymbol{c^*}, \epsilon} \epsilon$ subject to $C$
10 **if** $\epsilon^*_{new} \neq \epsilon^*$ **then**
11 $\epsilon^* := \epsilon^*_{new}$, go to line 4
12 **return** $(\boldsymbol{c^*}, \epsilon^*)$// *Best hyperplane and error*

Unfortunately, the drawback of this single linear programming approach is that for $M_{ij}$ linear constraints in $\mathit{Polytope}(\theta_{ij})$, the number of vertices of the polytope may be *exponential*, i.e., $N_{ij} = O(\exp M_{ij})$.

However, we make one final crucial insight: while we may have an exponential number of constraints in (4), we have a very efficient way to evaluate for a *fixed* solution $\boldsymbol{c}^*$, the single point $\boldsymbol{x^k_{ij}}$ in each $\mathit{Polytope}(\theta_{ij})$ with max error — this is exactly what the second stage linear program in (3) provides. Hence, this suggests an efficient *constraint generation* approach to solving (4) that we outline in Algorithm 2. Beginning with an empty constraint set, we iteratively add in constraints for the polytope vertices $\boldsymbol{x^k_{ij}}$ that yield maximal error for the current best solution $\boldsymbol{c^*}$ (one constraint for each side of the absolute value). Then we re-solve for $(\boldsymbol{c^*}, \epsilon^*)$ to see if the error has gotten worse; if not, we have reached optimality since $\boldsymbol{c^*}$ satisfies all constraints (vertices $\boldsymbol{x^k_{ij}}$ not having constraints in $C$ had provably equal or smaller error than those in $C$) and adding in all constraints could not reduce $\epsilon^*$ further.

We conclude our discussion of PairLeafApp in Algorithm 2 with a key observation: it will always terminate in finite time with the optimal solution, since at least two constraints are generated on every iteration and there are only a finite number of possible polytope vertices $\boldsymbol{x^k_{ij}}$ for which to generate constraints. We later demonstrate that PairLeafApp runs very efficiently in practice indicating that it is generating only a small subset of the possible exponential set of constraints.

# 4 Bounded Approximate Symbolic Dynamic Programming

Having shown how to efficiently approximate XADDs in Section 3, we switch to the main application focus of this work: finding bounded approximate solutions for Hybrid MDPs (HMDPs). Specifically, in this section, we build on the Symbolic Dynamic Programming (SDP) [14; 17] framework for HMDPs that uses the XADD data structure to maintain a compact representation of the value function, extending it to allow next-state dependent rewards and synchronic arcs in its transition function. In this work, we augment SDP with a bounded value approximation step that we will subsequently show permits the solution of HMDPs with strong error guarantees that cannot be efficiently solved exactly. We begin by formalizing an HMDP.

## 4.1 Hybrid Markov Decision Processes (HMDPs)

In HMDPs, states are represented by variable assignments. We assume a vector of variables $(\boldsymbol{b}^T, \boldsymbol{x}^T) = (b_1, \ldots, b_n, x_1, \ldots, x_m)$, where each $b_i \in \{0, 1\}$ $(1 \le i \le n)$ is boolean and each $x_j \in \mathbb{R}$ $(1 \le j \le m)$ is continuous. We also assume a finite set of $p$ parametrized actions $A = \{a_1(\boldsymbol{y}_1), \ldots, a_p(\boldsymbol{y}_p)\}$, where $\boldsymbol{y}_k \in \mathbb{R}^{|\boldsymbol{y}_k|}$ $(1 \le k \le p)$ denote continuous parameters for respective action $a_k$ (often we drop the subscript, e.g., $a(\boldsymbol{y})$).

An HMDP model also requires the following: (i) a joint state transition model $P(\boldsymbol{b}', \boldsymbol{x}'|\boldsymbol{b}, \boldsymbol{x}, a, \boldsymbol{y})$, which specifies the probability of the next state $(\boldsymbol{b}', \boldsymbol{x}')$ conditioned on a subset of the previous and next state and action $a(\boldsymbol{y})$; (ii) a reward function $R(\boldsymbol{b}, \boldsymbol{x}, a, \boldsymbol{y}, \boldsymbol{b}', \boldsymbol{x}')$, which specifies the immediate reward obtained by taking action $a(\boldsymbol{y})$ in state $(\boldsymbol{b}, \boldsymbol{x})$ and reaching state $(\boldsymbol{b}', \boldsymbol{x}')$; and (iii) a discount factor $\gamma$, $0 \le \gamma \le 1$.

A policy $\pi$ specifies the action $a(\boldsymbol{y}) = \pi(\boldsymbol{b}, \boldsymbol{x})$ to take in each state $(\boldsymbol{b}, \boldsymbol{x})$. Our goal is to find an optimal sequence of finite horizon-dependent policies $\Pi^* = (\pi^{*,1}, \ldots, \pi^{*,H})$ that maximizes the expected sum of discounted rewards over a horizon $h \in H; H \ge 0$:

$$V^{\Pi^*}(\boldsymbol{b}, \boldsymbol{x}) = E_{\Pi^*}\left[\sum_{h=0}^{H} \gamma^h \cdot r^h \middle| (\boldsymbol{b}_0, \boldsymbol{x}_0)\right]. \quad (5)$$

Here $r^h$ is the reward obtained at horizon $h$ following $\Pi^*$ where we assume starting state $(\boldsymbol{b}_0, \boldsymbol{x}_0)$ at $h = 0$.

HMDPs as defined above are naturally factored [3] in terms of state variables $(\boldsymbol{b}, \boldsymbol{x})$; as such, transition structure can be exploited in the form of a dynamic Bayes net (DBN) [6] where the conditional probabilities $P(b'_i|\cdots)$ and $P(x'_j|\cdots)$ for each next state variable can condition on the action, current and next

**Algorithm 3**: $\texttt{BASDP}(\text{HMDP}, H, \epsilon) \longrightarrow (V^h, \pi^{*,h})$

**begin**
  $V^0 := 0, h := 0$
  **while** $h < H$ **do**
    $h := h + 1$
    **foreach** $a \in A$ **do**
      $Q_a^h(\boldsymbol{y}) := \texttt{Regress}(V^{h-1}, a, \boldsymbol{y})$
      $Q_a^h := \max_{\boldsymbol{y}} Q_a^h(\boldsymbol{y})$ // *Parameter* max
      $V^h := \max_a Q_a^h$ // *Max all* $Q_a$
      $\pi^{*,h} := \arg\max_{(a,\boldsymbol{y})} Q_a^h(\boldsymbol{y})$
    $V^h = \texttt{XADDComp}(V^h, \epsilon)$
    **if** $V^h = V^{h-1}$ **then**
      break // *Stop if early convergence*
  **return** $(V^h, \pi^{*,h})$
**end**

**Algorithm 4**: $\texttt{Regress}(V, a, \boldsymbol{y}) \longrightarrow Q$

**begin**
  $Q = \texttt{Prime}(V)$ // *Rename all symbolic* //*variables* $b_i \to b_i'$ *and all* $x_i \to x_i'$
  $Q := R(\boldsymbol{b}, \boldsymbol{x}, a, \boldsymbol{y}, \boldsymbol{b'}, \boldsymbol{x'}) \oplus (\gamma \cdot Q)$
  // *Any var order with child before parent*
  **foreach** $v_k'$ *in* $Q$ **do**
    **if** $v_k' = x_j'$ **then**
      //*Continuous marginal integration*
      $Q := \int Q \otimes P(x_j' | \boldsymbol{b}, \boldsymbol{x}, \boldsymbol{v'}_{<k}, a, \boldsymbol{y})\, d_{x_j'}$
    **if** $v_k' = b_i'$ **then**
      // *Discrete marginal summation*
      $Q := \left[ Q \otimes P(b_i' | \boldsymbol{b}, \boldsymbol{x}, \boldsymbol{v'}_{<k}, a, \boldsymbol{y}) \right] |_{b_i'=1} \oplus \left[ Q \otimes P(b_i' | \boldsymbol{b}, \boldsymbol{x}, \boldsymbol{v'}_{<k}, a, \boldsymbol{y}) \right] |_{b_i'=0}$
  **return** $Q$
**end**

state. We allow *synchronic arcs* (variables that condition on each other in the same time slice) between any pair of variables, binary $\boldsymbol{b}$ or continuous $\boldsymbol{x}$ so long as they do not lead to cyclic dependencies in the DBN — this leads to a natural topologically sorted variable ordering that prevents any variable from conditioning on a later variable in the ordering. From these assumptions, we factorize the joint transition model as

$$P(\boldsymbol{b'}, \boldsymbol{x'} | \boldsymbol{b}, \boldsymbol{x}, a, \boldsymbol{y}) = \prod_{k=1}^{n+m} P(v_k' | \boldsymbol{b}, \boldsymbol{x}, \boldsymbol{v'}_{<k}, a, \boldsymbol{y})$$

where $\boldsymbol{v'}_{<k} = (v_1', \ldots, v_{k-1}'), 1 \le k \le n+m$.

The conditional probability functions $P(b_i' = v_{k_i}' | \boldsymbol{b}, \boldsymbol{x}, \boldsymbol{v'}_{<k_i}, a, \boldsymbol{y})$ for *binary* variables $b_i$ $(1 \le i \le n)$ can condition on state and action variables. For the *continuous* variables $x_j$ $(1 \le j \le m)$, we represent the CPFs $P(x_j' = v_{k_j}' | \boldsymbol{b}, \boldsymbol{x}, \boldsymbol{v'}_{<k_j}, a, \boldsymbol{y})$ with *piecewise linear equations* (PLEs) satisfying three properties: (i) PLEs can only condition on the action, current state, and previous state variables, (ii) PLEs are deterministic meaning that to be represented by probabilities they must be encoded using Dirac $\delta[\cdot]$ functions and (iii) PLEs are piecewise linear, where the piecewise conditions may be arbitrary logical combinations of the binary variables and linear inequalities over the continuous variables. Numerous examples of PLEs will be presented in the empirical results in Section 5.

While it is clear that our restrictions do not permit general stochastic continuous transition noise (e.g., Gaussian noise), they do permit discrete noise in the sense that $P(x_j' = v_{k_j}' | \boldsymbol{b}, \boldsymbol{x}, \boldsymbol{v'}_{<k_j}, a, \boldsymbol{y})$ may condition on $\boldsymbol{b'}$, which are stochastically sampled according to their CPFs. We note that this representation effectively allows modeling of continuous variable transitions as a mixture of $\delta$ functions, which has been used frequently in previous exact continuous state MDP solutions [7; 12].

We allow the reward function $R(\boldsymbol{b}, \boldsymbol{x}, a, \boldsymbol{y}, \boldsymbol{b'}, \boldsymbol{x'})$ to be a general piecewise linear function (boolean or linear conditions and linear values) such as

$$R(\boldsymbol{b}, \boldsymbol{x}, a, \boldsymbol{y}, \boldsymbol{b'}, \boldsymbol{x'}) = \begin{cases} b \wedge x_1 \le x_2 + 1 : & 1 - x_1' + 2x_2' \\ \neg b \vee x_1 > x_2 + 1 : & 3y + 2x_2 \end{cases}$$

The above transition and reward constraints ensure that all derived functions in the solution of these HMDPs will remain piecewise linear, which is essential for efficient linear XADD representation [14] and for the XADD approximation techniques proposed in Section 3.

### 4.2 Solution Methods

The algorithm we use for solving HMDPs is an approximate version of the continuous state and action generalization of *value iteration* [2], which is a dynamic programming algorithm for constructing optimal policies. It proceeds by constructing a series of $h$-stage-to-go optimal value functions $V^h(\boldsymbol{b}, \boldsymbol{x})$. Initializing $V^0(\boldsymbol{b}, \boldsymbol{x}) = 0$, we define the *quality* $Q_a^h(\boldsymbol{b}, \boldsymbol{x}, \boldsymbol{y})$ of taking action $a(\boldsymbol{y})$ in state $(\boldsymbol{b}, \boldsymbol{x})$ and acting so as to obtain $V^{h-1}(\boldsymbol{b}, \boldsymbol{x})$ thereafter as the following:

$$Q_a^h(\boldsymbol{b}, \boldsymbol{x}, \boldsymbol{y}) = \sum_{\boldsymbol{b'}} \int_{\boldsymbol{x'}} \Bigg[ \prod_{k=1}^{n+m} P(v_k' | \boldsymbol{b}, \boldsymbol{x}, \boldsymbol{v'}_{<k}, a, \boldsymbol{y}) \cdot \Big( R(\boldsymbol{b}, \boldsymbol{x}, a, \boldsymbol{y}, \boldsymbol{b'}, \boldsymbol{x'}) + \gamma V^{h-1}(\boldsymbol{b'}, \boldsymbol{x'}) \Big) \Bigg] d\boldsymbol{x'} \quad (6)$$

Given $Q_a^h(\boldsymbol{b}, \boldsymbol{x})$ for each $a \in A$, we can proceed to define the $h$-stage-to-go value function as the maximizing action parameter values $\boldsymbol{y}$ for the best action $a$ in each state $(\boldsymbol{b},\boldsymbol{x})$ as follows:

$$V^h(\boldsymbol{b},\boldsymbol{x}) = \max_{a\in A} \max_{\boldsymbol{y}\in\mathbb{R}^{|\boldsymbol{y}|}} \left\{ Q^h_a(\boldsymbol{b},\boldsymbol{x},\boldsymbol{y}) \right\} \qquad (7)$$

If the horizon $H$ is finite, then the optimal value function is obtained by computing $V^H(\boldsymbol{b},\boldsymbol{x})$ and the optimal horizon-dependent policy $\pi^{*,h}$ at each stage $h$ can be easily determined via $\pi^{*,h}(\boldsymbol{b},\boldsymbol{x}) = \arg\max_{(a,\boldsymbol{y})} Q^h_a(\boldsymbol{b},\boldsymbol{x},\boldsymbol{y})$. If the horizon $H = \infty$ and the optimal policy has finitely bounded value, then value iteration can terminate at horizon $h$ if $V^h = V^{h-1}$; then $V^\infty = V^h$ and $\pi^{*,\infty} = \pi^{*,h}$.

### 4.3 Bounded Approximate SDP (BASDP)

We will now define BASDP, our bounded approximate HMDP symbolic dynamic programming algorithm. BASDP is provided in Algorithm 3 along with a regression subroutine in Algorithm 4; BASDP is a modified version of SDP [17] to support the HMDP model with next-state dependent reward function and synchronic arcs as defined previously along with the crucial addition of line 10, which uses the `XADDComp` compression method described in Section 3. Error is cumulative over each horizon, so for example, the maximum possible error incurred in an *undiscounted* BASDP solution is $H\epsilon$. All functions are represented as XADDs, and we note that all of the XADD operations involved, namely addition $\oplus$, multiplication $\otimes$, integration of Dirac $\delta$ functions, marginalization of boolean variables $\sum_{b_i}$, continuous parameter maximization $\max_{\boldsymbol{y}}$ and discrete maximization $\max_a$, are defined for XADDs as given by [14; 17]. For most of these operations the execution time scales superlinearly with the number of partitions in the XADD, which can be greatly reduced by the `XADDComp` compression algorithm. We empirically demonstrate the benefits of approximation in the next section.

## 5 Empirical Results

In this section we wish to compare the scalability of exact SDP (calling BASDP in Algorithm 4 with $\epsilon = 0$) vs. various levels of approximation error $\epsilon > 0$ to determine the trade-offs between time and space vs. approximation error. To do this, we evaluated BASDP on three different domains — Mars Rover1D, Mars Rover2Dand Inventory Control— detailed next.

Mars Rover1D: A unidimensional continuous Mars Rover domain motivated by Bresina *et al* [5] used in order to visualize the value function and the effects of varying levels of approximation. The position of the rover is represented by a single continuous variable $x$ and the goal of the rover is to take pictures at specific positions. There is only one action $move(a_x)$, where $a_x$ is the movement distance. In the description of the problem for the instance shown below, there are two picture points and taking pictures is recorded in two boolean variables ($tp_1$ and $tp_2$). The dynamics for deterministic action $move(a_x)$ are as follows:

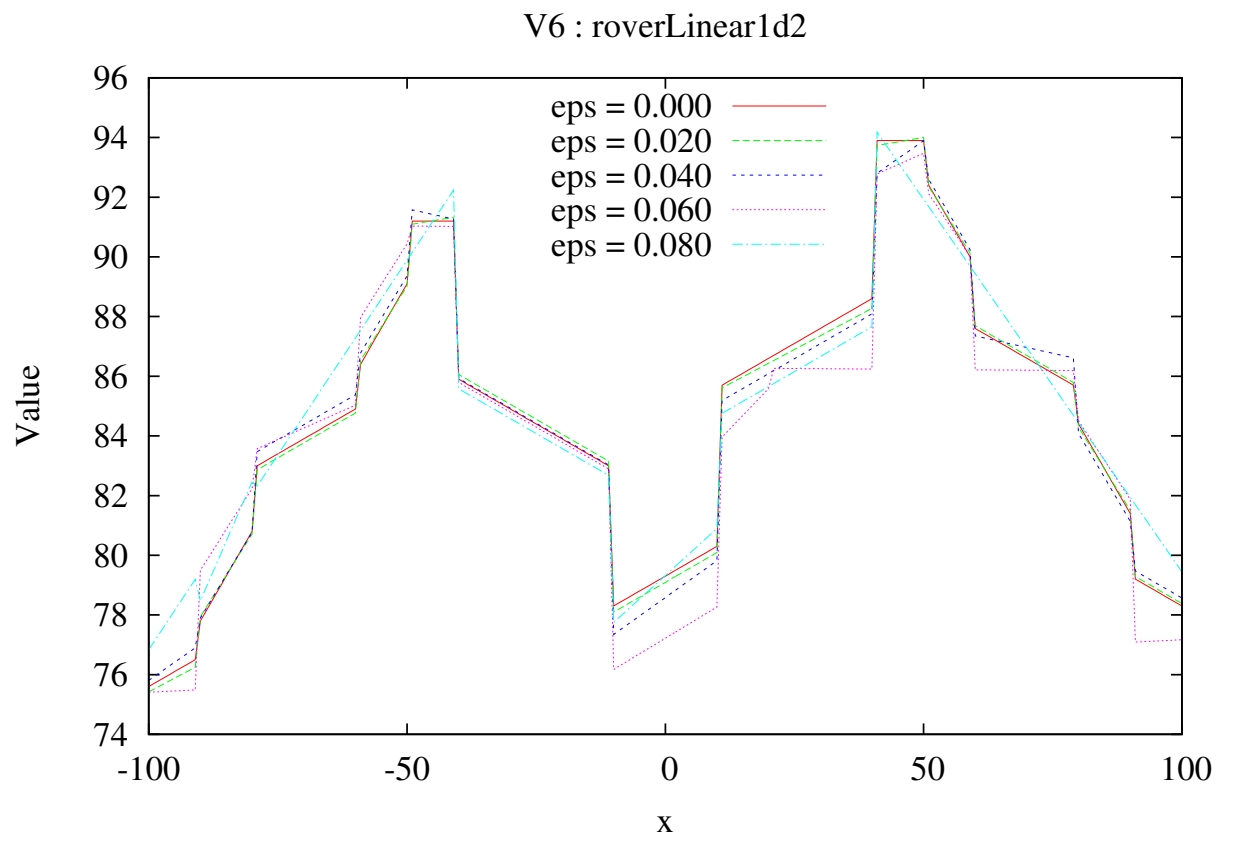

Figure 5: Value function at iteration 6 for Mars Rover1D, showing how different levels of approximation error (eps) lead to different compressions.

$$tp_1' = \begin{cases} tp_1 \vee (x > 40 \wedge x < 60) & : 1.0 \\ else & : 0.0 \end{cases}$$

$$tp_2' = \begin{cases} tp_2 \vee (x > -60 \wedge x < -40) & : 1.0 \\ else & : 0.0 \end{cases}$$

$$x' = x + a_x$$
$$R = R_1 + R_2 - 0.1 * |a_x|$$

$$R_1 = \begin{cases} (tp_1') \wedge (\neg tp_1) \wedge (x > 50) & : 40 - 0.2 * (x - 50) \\ (tp_1') \wedge (\neg tp_1) \wedge (x < 50) & : 40 - 0.2 * (50 - x) \\ (tp_1') \wedge (tp_1) & : 1.1 \\ else & : -2 \end{cases}$$

$$R_2 = \begin{cases} (tp_2') \wedge (\neg tp_2) \wedge (x > -50) & : 60 - 0.2 * (-x + 50) \\ (tp_2') \wedge (\neg tp_2) \wedge (x < -50) & : 60 - 0.2 * (x + 50) \\ (tp_2') \wedge (tp_2) & : 1.2 \\ else & : -1 \end{cases}$$

In Figure 5, we plot different value functions obtained by compressing with different levels — we note that in general larger $\epsilon$ results in a looser fit, but there are exceptions, owing to the greedy nature of successive pairwise merging for XADDs described in Section 3.

Mars Rover2D: In this multivariate version of a Mars Rover domain the rover is expected to follow a path. The position is represented by a pair of continuous variables $(x, y)$. There is only one action, $move(a_x, a_y)$, where $|a_x| < 10$ and $|a_y| < 10$. The new position is given by $(x', y') = (x + a_x, y + a_y)$. The reward increases with $x$ and decreases with the absolute value of $y$, that is:

$$R = \begin{cases} (x > y + 25) \wedge (x > -y + 25) \wedge (y > 0) : & -10 + x - y \\ (x > y + 25) \wedge (x > -y + 25) \wedge (y < 0) : & -10 + x + y \\ else : & -1 \end{cases}$$

In Figure 6, we can clearly see the effect of compression. In the 3D plots, a much simpler surface is obtained for the 5% error compression, and correspondingly, in the diagrams, the number of nodes is greatly reduced, which enables a much faster computation of XADD operations and the bounded error solution.

INVENTORY CONTROL: In an inventory problem [15], we assume $n$ continuous resources that can be bought and sold. There are $n$ *order-i* actions for each resource, $1 \leq i \leq n$. The maximum amount of each resource that is sold on one iteration depends on a stochastic demand variable $d$ that is true with 60% probability. The reward is equal to the sum of the resources sold in this iteration.The resource $x_i'$ for action *order-i* is given by:

$$x_i' = \begin{cases} (d') \wedge (x_i > 150) & : x_i + 200 - 150 \\ (d') \wedge (x_i < 150) & : 200 \\ (\neg d') \wedge (x_i > 50) & : x_i + 200 - 50 \\ (\neg d') \wedge (x_i < 50) & : 200 \end{cases}$$

and for other resources $x_j'$, $1 \leq j \leq n$, $j \neq i$:

$$x_j' = \begin{cases} (d') \wedge (x_j > 150) & : x_j - 150 \\ (d') \wedge (x_j < 150) & : 0 \\ (\neg d') \wedge (x_j > 50) & : x_j - 50 \\ (\neg d') \wedge (x_j < 50) & : 0 \end{cases}$$

Figure 7 shows the time, space, and actual error of the BASDP solutions vs. the exact solution for one MARS ROVER2D domain and one INVENTORY CONTROL domain. In the space plots (left), we note how the approximation compresses the XADD significantly, even for small $\epsilon$. We witness approximately 10× savings in time over the exact solution even for small $\epsilon$ and when we examine the actual error (right) of the BASDP solutions (compared to the exact solution), we see that it tends to be less than $\frac{1}{3}$ of the BASDP error bound.

# 6 Related Work

Boyan and Littman [4] presented the first exact solution for 1D continuous HMDPs with discrete actions, linear reward and piecewise dynamics while Feng *et al* [7] generalized this solution for a subset of multivariate HMDPs where all piecewise functions had to have rectilinear piece boundaries (i.e., general linear inequalities like $x + y > 0$ where disallowed) and actions were discrete. Li and Littman [10] extended Feng *et al*'s model to the case of bounded approximation using rectilinear piecewise *constant* functions that could not produce the low-error linear approximations in Figures 1(a,b) or 3. In addition, all of these methods could only provide (approximately) optimal solutions for a rectilinear subset of *discrete* action HMDPs in comparison to our more general setting of *continuous* action HMDPs with linear piecewise dynamics and rewards building on the work of Zamani *et al* [17].

An alternative bounded error HMDP solution is the phase-type approximation of Marecki *et al* [11] which can arbitrarily approximate 1D continuous MDP solutions but which does not extend to multivariate settings or continuous actions. In an approximate linear programming approach using basis functions, Kveton *et al* [9; 8] explore bounded approximations and learnable basis functions for HMDPs but cannot provide *a priori* guarantees on the maximum allowed error in a solution as we can in our BASDP framework. Munos and Moore [13] take a variable resolution discretization approach to refining value functions and policies, but these methods are based on rectilinear partitioned kd-trees which can consume prohibitive amounts of space to approximate the simple oblique piecewise linear function of Figure 2, represented exactly as a four node XADD.

In an orthogonal direction to the work above, Meuleau *et al* [12] investigate a search-based dynamic programming solution to HMDPs that is restricted to optimality over a subset of initial states. We note that this approach admits *any* dynamic programming backup and value representation and hence can be combined with BASDP and XADDs as proposed here — an interesting avenue for future work.

# 7 Concluding Remarks

In this work, we introduced a novel bounded approximate symbolic dynamic programming (BASDP) algorithm for HMDPs based on XADD approximation, where we contributed a bounded error compression technique for XADDs involving the solution of a constrained bilinear saddle point problem. After exploiting a number of key insights in the structure of this problem, we were able to show that it could be solved to optimality in finite time via constraint generation in a linear programming framework (despite having an apparent infinite set of potential constraints). Empirically, this BASDP solution yielded order-of-magnitude speedups over the exact solution in exchange for a small approximation error, thus vastly expanding the range of HMDPs for which bounded error approximate solutions are possible.

# Acknowledgments

NICTA is funded by the Australian Government as represented by the Department of Broadband, Communications and the Digital Economy and the Australian Research Council through the ICT Centre of Excellence program. This work was supported by the Brazillian agency FAPESP (grant 2012/10861-0).

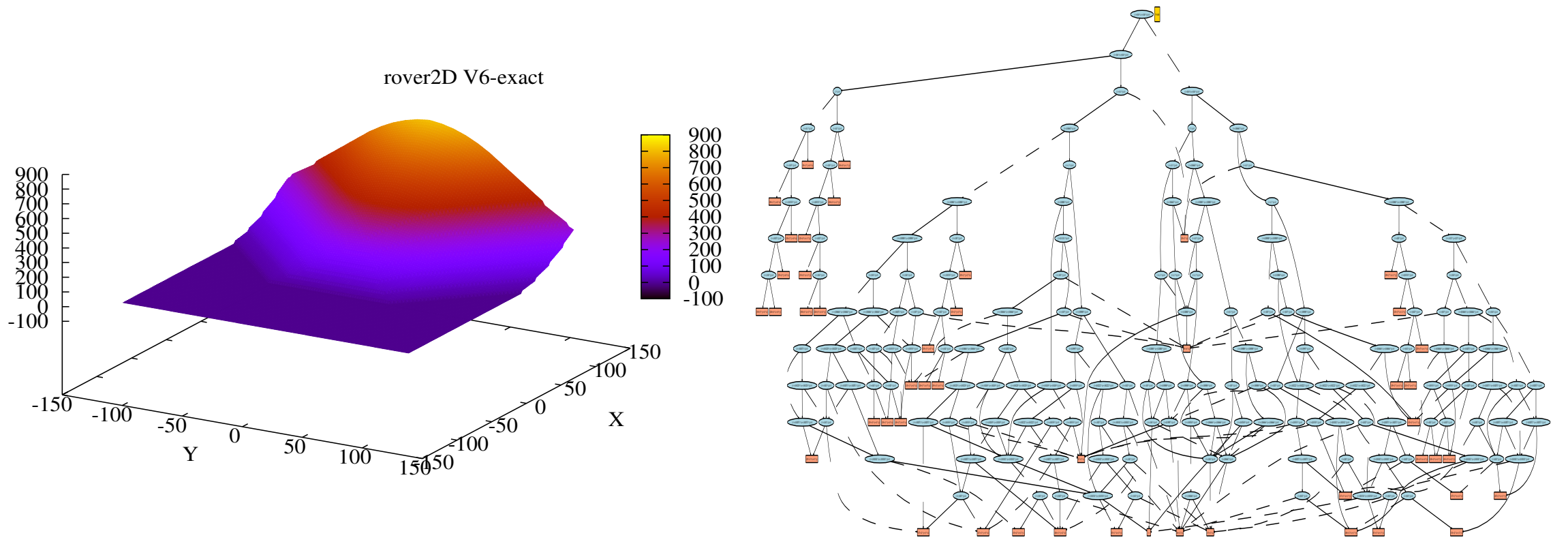

(a) Value at $6^{th}$ iteration for exact SDP.

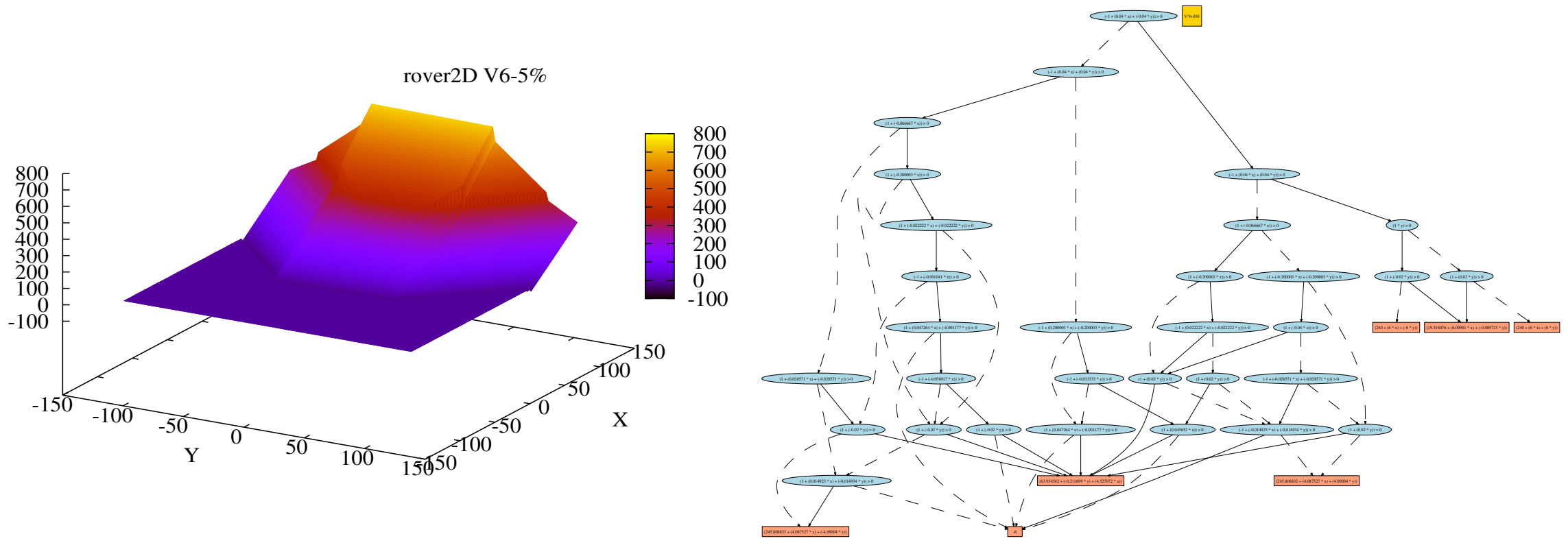

(b) Value at $6^{th}$ iteration for 5% approximate SDP.

Figure 6: Value function at iteration 6 for the Mars Rover2Ddomain; *(a)* Exact value function; *(b)* Approximate value function with error bounded 5% per iteration; *(left)* 3D Plots; *(right)* XADD Diagrams.

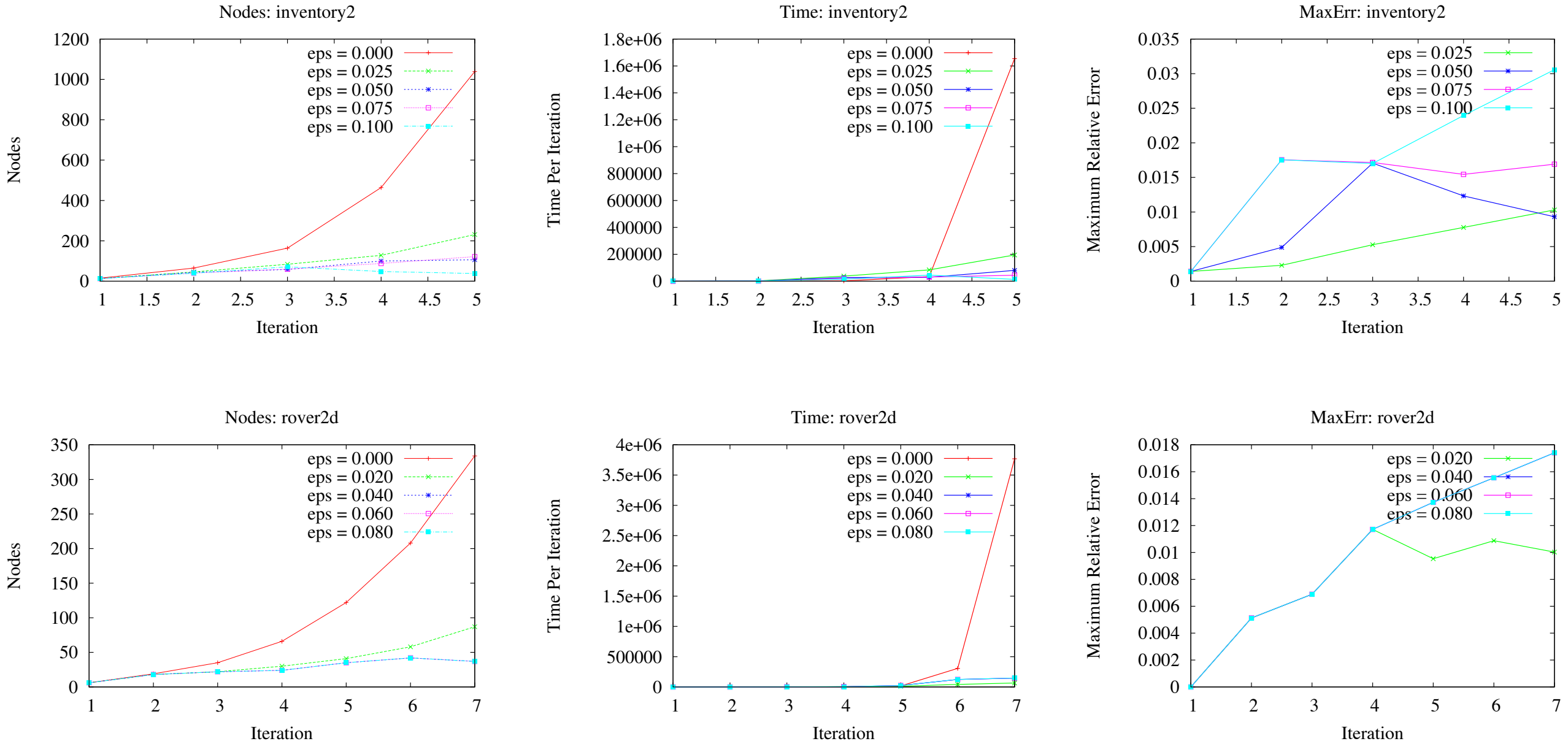

Figure 7: Performance plots for Mars Rover2D and Inventory Control2 with 5 different relative errors (eps): *(left)* Space (number of Nodes); *(middle)* Time (miliseconds); *(right)* Maximal error as fraction of the max value.